\documentclass{article}

    \usepackage[preprint]{neurips_2025}

\usepackage[utf8]{inputenc} 
\usepackage[T1]{fontenc}    
\usepackage{hyperref}       
\usepackage{url}            
\usepackage{booktabs}       
\usepackage{amsfonts}       
\usepackage{nicefrac}       
\usepackage{microtype}      
\usepackage{xcolor}         

\usepackage{graphicx}
\usepackage{amsmath}
\usepackage{makecell}

\title{Steering Risk Preferences in Large Language Models by Aligning Behavioral and Neural Representations}

\author{%
  Jian-Qiao Zhu\thanks{Equal contribution. Correspondence to: Jian-Qiao Zhu \texttt{<jz5204@princeton.edu>}}\\
  Department of Computer Science\\
  Princeton University\\
  \And
  Haijiang Yan$^*$ \\
  Department of Psychology \\
  University of Warwick \\
  \AND
  Thomas L. Griffiths \\
  Departments of Psychology and Computer Science \\
  Princeton University \\
}

\begin{document}

\maketitle

\begin{abstract}
Changing the behavior of large language models (LLMs) can be as straightforward as editing the Transformer’s residual streams using appropriately constructed ``steering vectors.'' These modifications to internal neural activations, a form of representation engineering, offer an effective and targeted means of influencing model behavior without retraining or fine-tuning the model. But how can such steering vectors be systematically identified? We propose a principled approach for uncovering steering vectors by aligning latent representations elicited through behavioral methods (specifically, Markov chain Monte Carlo with LLMs) with their neural counterparts. To evaluate this approach, we focus on extracting latent risk preferences from LLMs and steering their risk-related outputs using the aligned representations as steering vectors. We show that the resulting steering vectors successfully and reliably modulate LLM outputs in line with the targeted behavior.
\end{abstract}

\section{Introduction}

LLMs are increasingly deployed in risk-sensitive domains such as finance \cite{niszczota2023gpt} and healthcare \cite{abdou2025leveraging}. In these high-stakes applications, it is essential to develop reliable methods for aligning the behavior of LLMs with human values and safety requirements \cite{hendrycks2025introduction, mazeika2025utility}. One solution to this problem is {\em steering}, a term that refers to any targeted intervention (whether through model weights, decoding strategies, prompts, or internal neural activations) intended to control or shape a model’s outputs. Steering risk-related behavior in LLMs is one way to ensure alignment with humans in risky domains.

Steering the risk preferences of a pretrained LLM, however, is inherently challenging due to the opacity of these models. LLMs operate over vast weight spaces, and their outputs are highly context-dependent \cite{brown2020language, zhu2024incoherent}, making it difficult to isolate or manipulate any specific internal variable that governs risk preference. Existing steering techniques, such as prompt engineering or supervised fine-tuning, either lack the granularity needed to target specific latent representation or require extensive retraining and human supervision \cite{qi2023fine, ziegler2019fine}.

Given the context-dependent and probabilistic nature of LLM outputs, we propose that their underlying risk preferences are best characterized as \textit{probabilistic representations}. That is, the same risky decision may yield different completions depending on subtle variations in input phrasing or prompt structure \cite{zhu2024incoherent, brown2020language}. This view suggests that an LLM’s underlying risk preferences can be recovered by sampling its behavior over many such comparisons. To operationalize this idea, we measure the risk preferences of LLMs using a method based on Markov chain Monte Carlo (MCMC; \cite{sanborn2007markov}), using repeated choices by the model to define a Markov chain that converges to a probability distribution that captures these preferences. This approach has previously been used to elicit probabilistic representations in other settings from both humans and LLMs \cite{noguchi2013non, harrison2020gibbs, zhu2024recovering}.

Measuring the risk preferences of LLMs in this way creates the opportunity to build a bridge between the observed behavior and the activations of nodes in the underlying neural network \cite{sucholutsky2023getting}. We can  align the behavioral representations we find with the neural representations within LLMs and use this alignment to derive a steering vector that captures underlying risk preferences. When this vector is injected back into the model at inference time, it enables precise control of the model’s risk-related behavior. We refer to this approach as {\em self-alignment}, since it leverages the model's own emergent representations for behavior control.

To evaluate this method, we apply steering vectors derived from the aligned representation across three domains: risky decision-making, risk perception, and risk-related text generation. Our results demonstrate that self-alignment yields substantially greater control over model behavior than two other baselines: (i) the Contrastive Activation approach, which derives steering vectors from prompt pairs, and (ii) a Certainty Equivalent control condition that uses an alternative behavioral elicitation. Our results also demonstrate that the modified risk preferences transfer to tasks that are quite far away from the choices from which the steering vectors were derived.

\section{Background}

\textbf{AI safety and value alignment.}
AI safety research has long emphasized the importance of aligning artificial systems with human values, though explicitly encoding such values in machines remains a formidable challenge \cite{russell2022human}. The emergence of LLMs presents new opportunities in this regard, as LLMs have been shown to internalize commonsense knowledge and human norms from large-scale training data \cite{hendrycks2020aligning, mazeika2025utility, marjieh2024large}. Consequently, many researchers now argue that, given sufficient data, LLMs can approximate shared norms \cite{brown2020language}. At the same time, however, this capacity introduces new challenges for interpretability, as it becomes increasingly difficult to discern the underlying factors that drive LLM behavior. In this work, we propose a novel strategy that leverages emergent representations in LLMs by eliciting the same latent construct through both behavioral and neural means. This dual elicitation facilitates self-alignment between behavior and neural activations, providing an interpretable method for steering LLM outputs.


\textbf{LLM steering.}
A range of approaches has been proposed to influence the outputs of pretrained LLMs, which can broadly be categorized as different forms of intervention. These interventions vary in where and how they modify the model. Weight-level interventions include techniques such as supervised fine-tuning \cite{qi2023fine} and reinforcement learning with human feedback \cite{ziegler2019fine}, which directly update model parameters. Alternatively, decoding-level interventions, such as trainable decoding, modify the output generation process while keeping model weights fixed \cite{grover2019bias}. Prompt engineering can be viewed as an intervention on the input space, shaping model behavior through carefully constructed prompts \cite{zhou2022steering, yao2023tree}. Moreover, activation-level interventions, which typically freeze the model weights and instead search for steering vectors, offer an alternative to behavioral control. These vectors can be discovered through gradient-based optimization \cite{hernandez2023inspecting} or computed directly from contrastive prompt pairs \cite{li2023inference, turner2023steering}. In this work, we propose an alignment-based method for deriving steering vectors by aligning behavioral and neural representations of latent constructs such as risk preference.

\section{Method}

\begin{figure}[t!]
    \centering
    \includegraphics[width=0.9\linewidth]{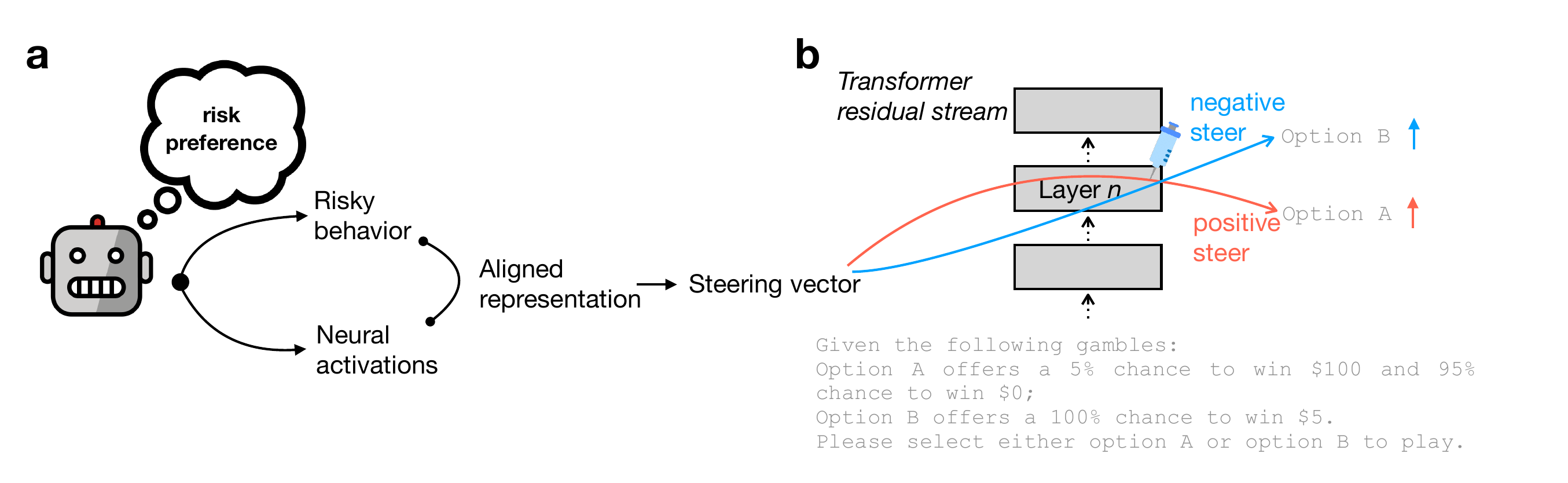}
    \caption{{\bf Aligned steering vectors.} \textbf{(a)} Overview of the proposed method for generating steering vectors by aligning representations of risk preference derived from behavioral and neural elicitation. \textbf{(b)} During inference, the steering vector is injected into the residual stream at all token positions to control LLM outputs. When the steering vector is applied with a positive multiplier (i.e., positive steering), the LLM is expected to exhibit more risk-seeking behavior. Conversely, applying a negative multiplier (i.e., negative steering) is expected to induce more risk-averse behavior. }
    \label{fig:method_illustrate}
\end{figure}

The key idea behind our proposed method is to derive a steering vector that optimally aligns the model's behavioral and neural representations of risk preference (see Figure \ref{fig:method_illustrate}a). Our method proceeds in two main steps, described below.

\textbf{Step 1: Eliciting behavioral representations of risk via MCMC.}
Risk preference, like many other mental representations, is inherently unobservable. However, as demonstrated in cognitive psychology, such latent constructs can be inferred from observed behavior \cite{kay2023world, sanborn2007markov, harrison2020gibbs}. Drawing inspiration from recent work on behavioral elicitation in LLMs \cite{zhu2024eliciting, zhu2024recovering, capstick2024using}, we incorporate LLMs into a MCMC sampler to effectively elicit their behavioral representation of risk. A variety of sampling algorithms can be used for this purpose, including the Metropolis-Hastings algorithm \cite{zhu2024recovering, sanborn2007markov} and Gibbs sampling \cite{harrison2020gibbs}. The core intuition is to use the LLM to define the proposal or acceptance mechanism, such that the resulting sequence of samples produced by the Markov chain converges to a stationary distribution that reflects the model's latent representations \cite{zhu2024recovering, noguchi2013non, harrison2020gibbs, leon2020uncovering, yan2024quickly, sanborn2007markov}.

\begin{figure}[t!]
    \centering
    \includegraphics[width=0.9\linewidth]{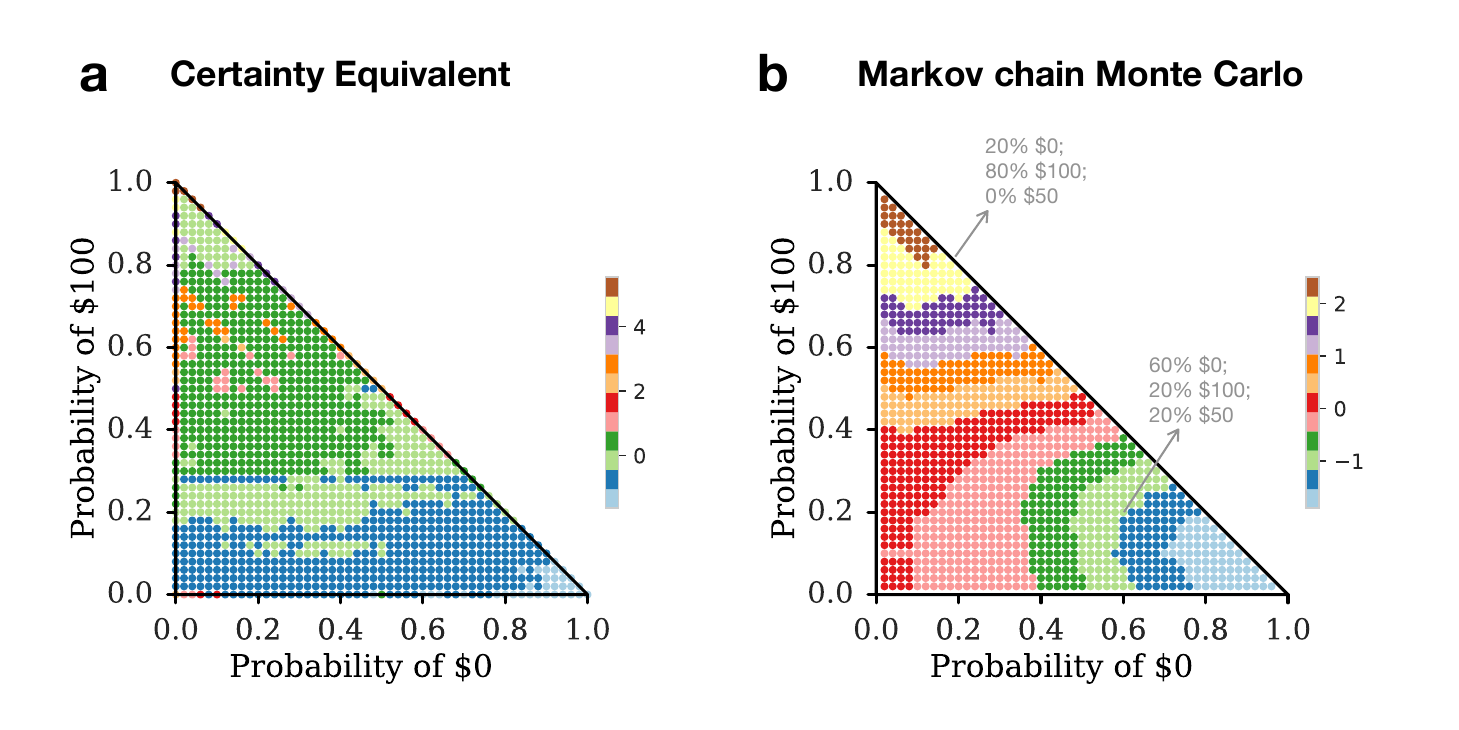}
    \caption{\textbf{Elicited risk preferences from Gemma-2-9B-Instruct using behavioral methods.} \textbf{(a)} Certainty Equivalent method. \textbf{(b)} Markov chain Monte Carlo with LLM. Each triangle represents the probability simplex over three-outcome gambles (\$0, \$50, and \$100), where the sum of outcome probabilities equals one. The MCMC-with-LLM elicitation reveals more nuanced and structured contours of risk preference compared to the Certainty Equivalent method. Higher values indicate a stronger preference for the gamble by the Gemma model.}
    \label{fig:behavior_risk}
\end{figure}

Specifically, we adapted the procedure established by \cite{noguchi2013non}, which has been successfully used to elicit human risk preferences, to the LLM setting. In each trial, the LLM is prompted to choose between two gambles, A and B. In our implementation, all gambles consist of three possible outcomes: \$0, \$50, and \$100. While the outcome values are fixed, the probabilities associated with each outcome vary across gambles. After the LLM makes a choice, the selected gamble is retained, and the unchosen option is replaced with a newly generated gamble. The probabilities for this new gamble are randomly sampled from a Dirichlet distribution: $\text{Dir}(1,1,1)$. In the next trial, the retained and newly generated gambles are presented again (with their order randomized), and the LLM is asked to make another choice. Importantly, no choice history is provided in the prompt: each decision is made solely based on the current pair of gambles presented. 
The sequence of choices made by the LLM forms the foundation for constructing its behavioral representation of risk. Specifically, within the space of all possible gambles, represented as a probability triangle\footnote{In the economics literature, this triangle is also known as the Marschak–Machina probability triangle \cite{marschak1950rational, machina1982expected}, a method traditionally used to qualitatively differentiate among competing theories of risky choice \cite{wu1998common}. Our MCMC approach provides a non-parametric estimation of the LLM’s risk representation within this triangle.}, the LLM’s choices allow us to infer a probability distribution over gambles reflecting its preferences across risk profiles (see Figure \ref{fig:behavior_risk}).

More formally, MCMC begins at an initial state $z$ (i.e., a specific gamble from the probability triangle). A proposed next state $z'$ is drawn from a proposal distribution $q(z'|z)$, and is then evaluated under the target distribution $\pi$ (i.e., latent representation of risk) to determine whether it should be accepted as the new state or rejected in favor of retaining the current state $z$. To guarantee that the Markov chain converges to $\pi$, it is sufficient to satisfy the condition of detailed balance (along with ergodicity):
\begin{align}
    \pi(z)q(z'|z)A(z',z) = \pi(z')q(z|z')A(z,z')
\end{align}
where $q(z'|z)$ is the probability of proposing $z'$ from state $z$, and $A(z',z)$ is the probability of accepting proposal $z'$ over $z$. In our case, we use a symmetric proposal distribution (i.e., $q(z'|z)=q(z|z')$), which simplifies the detailed balance condition to: $\pi(z)A(z',z) = \pi(z')A(z,z')$.
One way to satisfy this condition is by using the Barker acceptance function \cite{barker1965monte}: 
\begin{align}
    A(z',z) = \frac{\pi(z')}{\pi(z)+\pi(z')}
\end{align}
This acceptance rule is particularly appropriate for modeling LLM behavior, as it closely resembles well-known stochastic choice models such as Luce’s choice rule \cite{luce1959individual} and the Bradley–Terry model \cite{rafailov2023direct}, which has been applied in LLM post-training and preference alignment. 
As a result, by sequentially presenting pairs of risky choice alternatives to an LLM, the set of selected options can be interpreted as samples from a probability distribution whose density is proportional to the LLM’s latent representation of risk \cite{noguchi2013non}.

\textbf{Step 2: Aligning behavioral and neural representations to compute steering vectors.}
Given the behavioral representations of risk elicited via MCMC with LLM, we next sought to align them with the model’s internal neural activations. To obtain corresponding neural representations of risk preference for each gamble, we prompted the same LLM to evaluate the attractiveness of the gamble when hypothetically offered (see Appendix \ref{ap:prompts_for_neural_acts} for details). We then aligned the two representations by regressing the behavioral estimates onto the neural activations. Specifically, we treated the neural activations as independent variables and the behavioral responses as dependent variables, using Lasso regression with an L1 penalty of 10. The resulting regression coefficients, corresponding to neurons in the Transformer’s residual stream, are interpreted as reflecting the LLM’s risk preference and are thus used as the steering vector.

\section{Other Methods}

\textbf{Contrastive Activation.} 
Another simple yet effective method for computing steering vectors is to contrast intermediate neural activations on carefully selected prompt pairs--a technique known as \textit{Contrastive Activation} \cite{panickssery2023steering, turner2023steering}. For example, to steer LLM outputs toward more positive sentiment, Contrastive Activation compares the model’s internal activations on a contrasting pair of prompts such as ``Love'' and ``Hate'' \cite{turner2023steering}. The difference between these activations is treated as the steering vector, which is then added to the model’s residual stream during inference. This shifts the model’s internal representations along the desired semantic direction (e.g., toward ``Love'' and away from ``Hate''), resulting in completions that reflect more positive sentiment.

To adapt Contrastive Activation to the task of steering LLM risk preferences, we constructed a list of words associated with “Risk” and “Safety” (see Appendix \ref{ap:CA_words} for details). Following the standard procedure for computing steering vectors in Contrastive Activation, we extracted the residual stream activations of the LLM for each word across multiple layers. The steering vector was computed as the average difference in residual activations between the risk-related and safety-related word pairs.

\textbf{Certainty Equivalent.}
Finally, we consider an alternative behavioral method for eliciting individuals’ risk preferences that is widely used in economics and psychology: the \textit{Certainty Equivalent}. The Certainty Equivalent refers to the sure amount of money a person is willing to accept in place of a risky gamble \cite{von1947theory, kahneman1979prospect}. In other words, it represents the value at which an individual is indifferent between receiving a certain payoff or a probabilistic outcome. This measure serves as a behavioral proxy for risk preference: individuals are classified as risk-averse if their Certainty Equivalent is lower than the gamble’s expected value, risk-neutral if it is equal, and risk-seeking if it is higher.

In our task, the Certainty Equivalent serves as a direct control condition for the risk representations elicited via MCMC. Specifically, we elicited the LLM’s Certainty Equivalent for all gambles previously used in the MCMC procedure. This produces an alternative behavioral representation of risk, based on Certainty Equivalents rather than MCMC, while holding the set of gambles constant. 

\section{Experiments}

In this work, we focus on steering the risk preferences of Gemma-2-9B-Instruct \cite{team2024gemma}, which serves as our primary target LLM. We also replicate the main experiments using the smaller-scale Gemma-2-2B-Instruct (see Appendix \ref{ap:replication}). The temperature was fixed at 1 for behavioral elicitation.

\textbf{Behavioral elicitation of risk representations.}
For both the MCMC with LLM and Certainty Equivalent methods, we derive steering vectors from behaviorally aligned representations of risk. That is, these approaches rely on first eliciting the model’s risk preferences through behavioral methods. To obtain a quantitative characterization of these preferences, we focus on the space of gambles defined over the probability triangle (see Figure~\ref{fig:behavior_risk}).

In the Certainty Equivalent method, we probed Gemma-2-9B-Instruct by densely sampling gambles across the probability triangle. For each gamble, the model was prompted to report its certainty equivalent. These responses were then aggregated and normalized across all sampled gambles to produce the density plot shown in Figure \ref{fig:behavior_risk}a.

Similarly, for the MCMC with LLM method, we embedded the Gemma model within a MCMC sampler, prompting it to accept or reject newly proposed gambles through binary choices. The space of gambles was identical to that used in the Certainty Equivalent method (i.e., the probability triangle). The Markov chain consisted of 3,000 such binary choices. The resulting risk representation elicited via MCMC (smoothed using a Dirichlet kernel of width 0.09 that preserves probability triangle boundaries) is shown in Figure \ref{fig:behavior_risk}b. Note that the behavioral representation derived from MCMC reveals more nuanced gradients in the density plot, highlighting a stark contrast with the coarser structure observed in the Certainty Equivalent method. 

Steering vectors for both the MCMC and Certainty Equivalent methods were computed using Lasso regression to align behavioral and neural representations (see Appendix \ref{ap:compare_steer_vector} for a comparison). In contrast, the steering vector for the Contrastive Activation method was derived by computing the difference in neural activations between pairs of risk-related and safety-related words. Steering vectors were normalized by division by their Euclidean norm.

\begin{table}[h!]
    \centering
    \caption{Gambles used to evaluate the effectiveness of steering LLMs’ risky decision-making. Risky options are expressed in the format \{probability, outcome\}; the remaining probability corresponds to receiving nothing.  }
    \begin{tabular}{lcc} \toprule
         & \multicolumn{2}{c}{Outcome probability} \\ \cline{2-3}
            & High & Low \\ \hline
         Gains & \{80\%, \$4000\} vs \$3000 & \{5\%, \$100\} vs \$5 \\
         Losses & \{80\%, -\$4000\} vs -\$3000 & \{5\%, -\$100\} vs -\$5 \\ \bottomrule
    \end{tabular}
    \label{tab:four_fold_gambles}
\end{table}

\textbf{Steering LLM risky choices.} 
Having obtained three steering vectors (derived from MCMC with LLM, Contrastive Activation, and Certainty Equivalent methods), we now evaluate the effectiveness of the three steering vectors in controlling LLM’s risky decision-making. Our analysis focuses on a set of four gambles that have been foundational in the study of the fourfold pattern of risk preferences \cite{kahneman1979prospect}. This well-documented pattern describes how human decision-makers tend to be risk-seeking when the probability of a positive outcome is low and when the probability of a negative outcome is high; conversely, they tend to be risk-averse when the probability of a positive outcome is high and when the probability of a negative outcome is low (see Table \ref{tab:four_fold_gambles} for examples).

To steer the LLM’s decisions toward greater risk-seeking or risk-aversion, we prompted the Gemma model with the gambles shown in Table \ref{tab:four_fold_gambles}, framed as binary choices between options A and B. During inference, steering vectors (scaled by a predefined multiplier ranging between -900 to +900) were added to the model’s residual stream at each token position. The model then continued its forward pass to the output layer, where we extracted the token probabilities for ``A'' and ``B'' from the final logits. These probabilities were normalized and then used to quantify the model's choice behavior under different steering conditions. 

\begin{figure}[t!]
    \centering
    \includegraphics[width=0.95\linewidth]{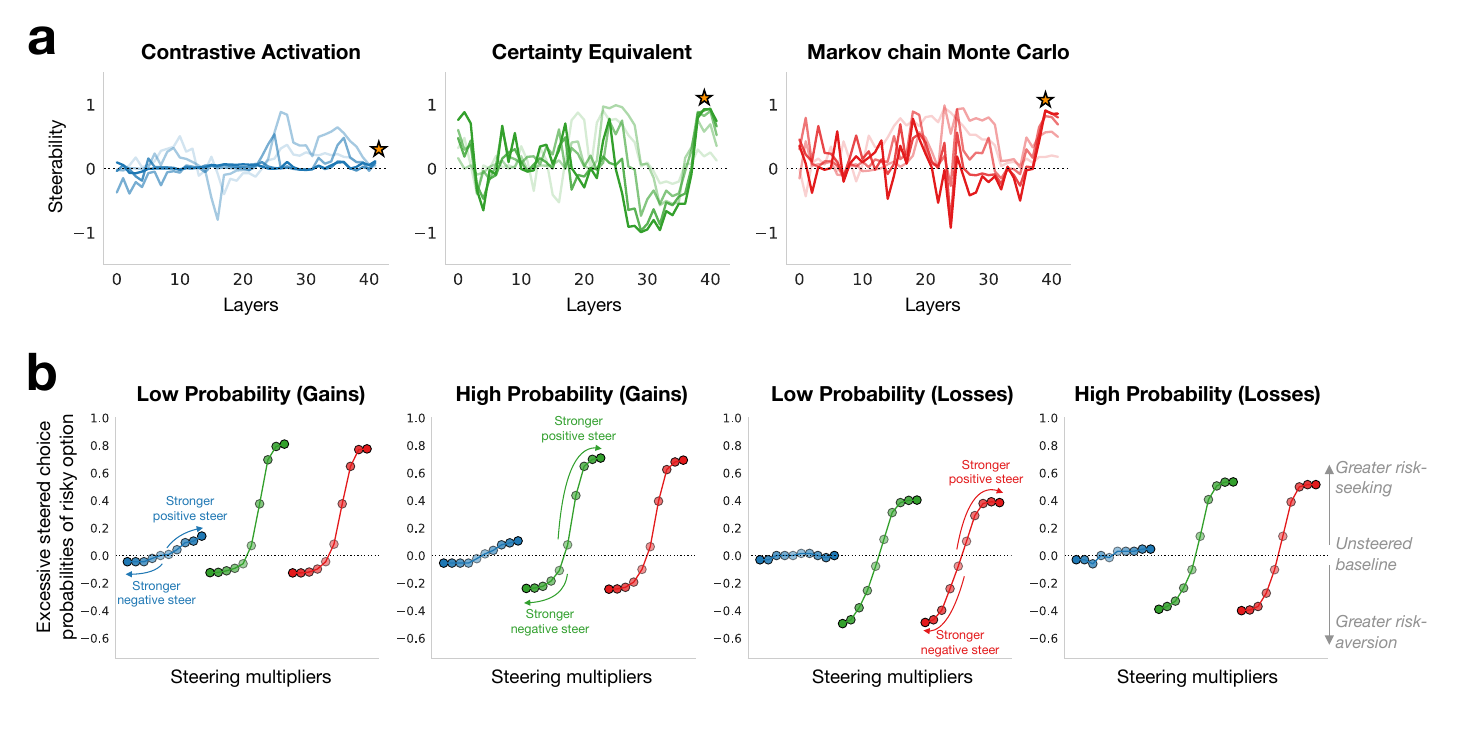}
    \caption{\textbf{Steering risky decisions of Gemma-2-9B-Instruct.} \textbf{(a)} Steerability results using steering vectors derived from Contrastive Activation (blue), Certainty Equivalent (green), and MCMC (red). Darker colors indicate larger steering multipliers. The optimal layers for steering, identified by the highest steerability at the maximum multiplier, are layers 41, 39, and 39 for the three methods, respectively (marked with stars). \textbf{(b)} Change in choice probabilities for the risky option after steering, using the best layer for each method. The vertical axis reflects the difference from the unsteered baseline probabilities across the four gambles.  }
    \label{fig:steer_risk_decisions}
\end{figure}

We first evaluated which Transformer layer contains the most effective residual stream for steering. To do so, we examined the model’s behavior under extreme steering conditions, using multipliers of –900 and +900. For each layer, we computed the steered choice probabilities with each steering multiplier. We then quantified steerability as the average difference in choice probabilities across four gambles (see Figure \ref{fig:steer_risk_decisions}a), defined as:
\begin{align}
    \text{Steerability} = \frac{1}{4}\sum_{i=1}^4 \Big( p_\text{positive}(z_i) - p_\text{negative}(z_i) \Big)
\end{align}
where $z_i$ denotes the $i$-th gamble prompted to the LLM, $p_\text{positive}(z_i)$ is the model's probability of choosing the risky option under positive steering, and $p_\text{negative}(z_i)$ is the corresponding probability under negative steering of equal magnitude.

As shown in Figure \ref{fig:steer_risk_decisions}b, we compared the steered choice probabilities for the risky option by subtracting the baseline (unsteered) choice probabilities. A value of zero therefore indicates no change relative to the unsteered baseline. We find that the steering vector derived from the Contrastive Activation method has limited impact on altering the Gemma model’s choice behavior. In contrast, steering vectors computed by aligning behavioral and neural representations (i.e., both Certainty Equivalent and MCMC methods) are effective in controlling the Gemma model’s risky decision-making, shifting it toward greater risk-seeking under positive steering and greater risk aversion under negative steering.

\textbf{Steering LLM risk perception.}
While studying abstract gambles provides valuable insights into an agent's risk preferences, psychologists have also used more naturalistic stimuli to assess people's perception of risk in real-world contexts such as ``cheating on an exam'' or ``forging someone’s signature'' \cite{weber2002domain, slovic1987perception}. LLMs, by virtue of their broad training data, are presumably capable of forming meaningful risk perceptions about such real-world events \cite{mazeika2025utility, turner2023steering}. Indeed, recent work has shown that LLM embeddings account for a substantial portion of the variance in human risk perception \cite{bhatia2024exploring}. Here, we investigate the extent to which an LLM’s risk perception can be steered using the same set of steering vectors derived in the preceding analyses.

We prompted Gemma-2-9B-Instruct to rate real-world risky events using integers from 1 (not risky at all) to 7 (extremely risky). The full set included 150 risky events curated by \cite{bhatia2024exploring}, spanning a range of domains: ethical (e.g., ``passing off somebody else's work as your own''), financial (e.g., ``betting at the horse races''), health-related (e.g., ``consuming excessive amounts of alcohol''), sports (e.g., ``bungee jumping''), and social (e.g., ``trusting a stranger with your personal information''). As in previous experiments, we modified the residual stream during inference by injecting the steering vectors. However, instead of focusing on choices between options A and B, we extracted the model’s output logits for the integer tokens ``1'' through ``7.'' These token probabilities were then normalized to yield a distribution reflecting the model’s perceived risk level for each event.

\begin{figure}[t!]
    \centering
    \includegraphics[width=0.9\linewidth]{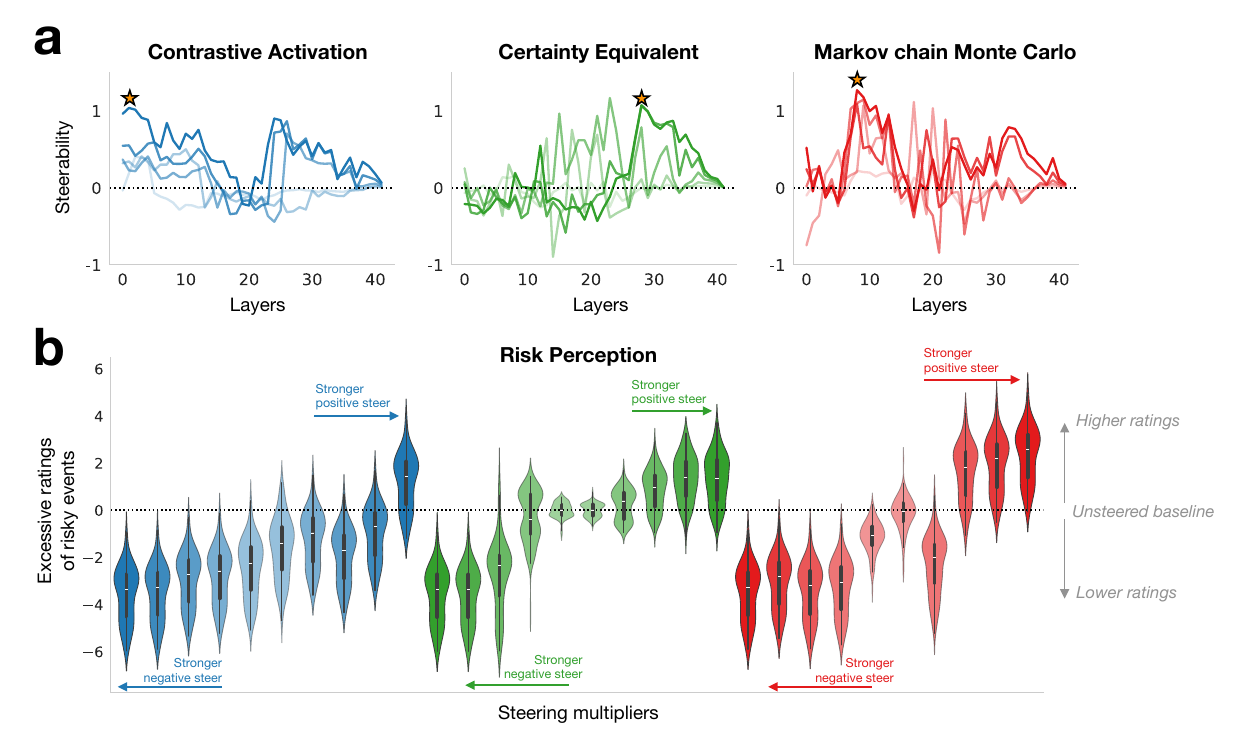}
    \caption{\textbf{Steering risk perception of Gemma-2-9B-Instruct.} \textbf{(a)} Steerability results using steering vectors derived from Contrastive Activation (blue), Certainty Equivalent (green), and MCMC (red). Darker colors represent larger steering multipliers. The optimal layers for steering, identified by the highest steerability at the maximum multiplier, are layers 2, 28, and 8 for the respective methods (marked with stars). \textbf{(b)} Change in average risk ratings for real-world events after steering, using the optimal layer for each method. The vertical axis reflects the deviation from the unsteered baseline rating. Each violin plot displays the distribution of ratings, with the white bar indicating the median and the black box representing the interquartile range up to the 75th percentile. }
    \label{fig:steer_risk_perception}
\end{figure}

Analogous to the steerability metric used for risky decisions, we define steerability for risk perception as the average difference in risk ratings across all real-world events between positive and negative steering conditions. However, by comparing the most steerable layers for risky decisions (Figure \ref{fig:steer_risk_decisions}a) and risk perceptions (Figure \ref{fig:steer_risk_perception}a), we find that risk perceptions are more effectively influenced at earlier layers, whereas risky decisions are more steerable in later layers closer to the output of the Gemma model. This finding aligns with psychological evidence suggesting that perceptual processes occur earlier than decision-making processes in the cognitive hierarchy \cite{vanrullen2001time}.

To evaluate the influence of steering vectors on model-generated ratings, we conducted two separate two-way repeated-measures ANOVAs for positive and negative steering conditions. For positive steering, the analysis revealed a significant main effect of steering method, $F(2, 298)=1717.09, p<.01$, a significant main effect of steering multiplier, $F(4, 596)=2503.30, p<.01$, and a significant interaction between method and multiplier, $F(8, 1192)=1507.64, p<.01$. Follow-up paired t-tests comparing steered ratings showed that the MCMC method significantly outperformed both the Certainty Equivalent method ($t(149)=2.41, p<.01$) and the Contrastive Activation method ($t(149)=88.95, p<.01$). Additionally, the Certainty Equivalent method significantly outperformed the Contrastive Activation method ($t(149)=38.90, p<.01$). 

Similarly, for negative steering, the two-way repeated-measures ANOVA revealed significant main effects of steering method ($F(2,298)=265.04, p<.01$), steering multiplier ($F(4,596)=832.89, p<.01$), and their interaction ($F(8,1192)=283.95, p<.01$). Follow-up paired t-tests showed that the MCMC method significantly outperformed both the Certainty Equivalent method ($t(149)=16.07, p<.01$) and the Contrastive Activation method ($t(149)=11.40, p<.01$). Moreover, the Contrastive Activation method yielded significantly better performance than the Certainty Equivalent method ($t(149)=16.70, p<.01$).

\textbf{Steering text generation for risky events in LLM.}
Finally, beyond quantitative evaluations based on choice probabilities and risk ratings, we also examine whether modifying the final layer's residual stream at inference time systematically alters the textual outputs of the Gemma model. Using the same set of 150 real-world risky events described above \cite{bhatia2024exploring}, we prompted the model with the sentence ``I think \{event\}'', where \{event\} is replaced with each risky scenario (e.g., ``I think cheating on an exam \_\_\_'' or ``I think consuming excessive amounts of alcohol \_\_\_''). Steering vectors were injected into the final layer's residual stream at each subsequent token position during generation. 

\begin{figure}[t!]
    \centering
    \includegraphics[width=\linewidth]{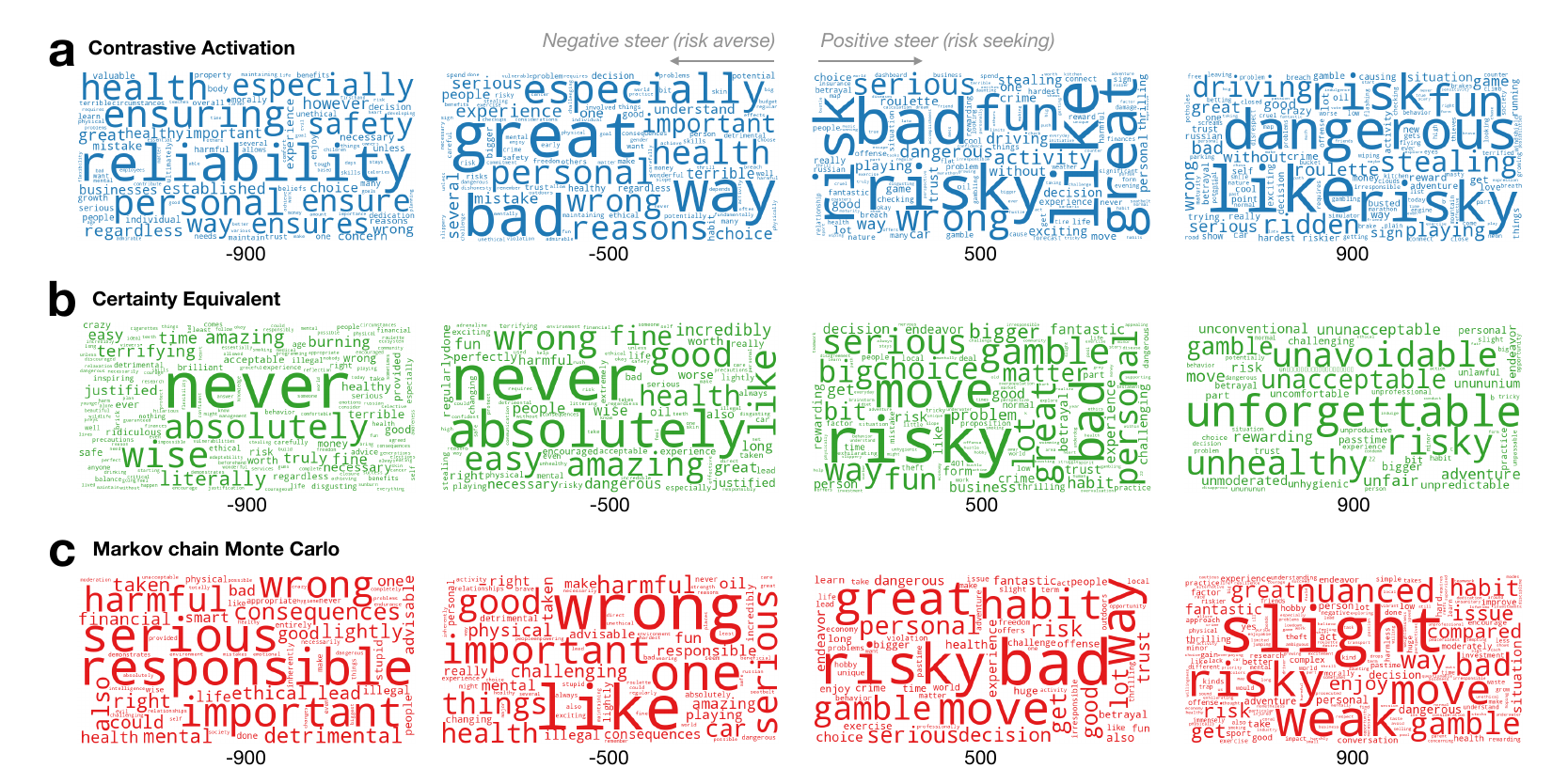}
    \caption{\textbf{Steering text generation for real-world risky events in Gemma-2-9B-Instruct.} Text outputs generated by injecting steering vectors into the residual stream at the final layer of the model during inference. Steering vectors are derived from \textbf{(a)} Contrastive Activation, \textbf{(b)} Certainty Equivalent, and \textbf{(c)} MCMC with LLM methods. Each word cloud represents the frequency distribution of words used in the model’s completions under different steering conditions. The corresponding steering multiplier is indicated below each word cloud.}
    \label{fig:steer_risk_language}
\end{figure}

To give a visual idea of how steering influences model-generated text, we created word clouds that reflect the frequency of word usage following the application of steering vectors (see Figure \ref{fig:steer_risk_language}). Overall, the Gemma model recognizes the inherent risk in the real-world events presented. However, when the residual stream at the final layer is positively steered toward risk-seeking behavior using the vector derived from the MCMC method, we observe a noticeable attenuation in the model's perceived risk. For example, completions more frequently include phrases such as ``slightly risky'' and ``a minor offense'' (see Figure \ref{fig:steer_risk_language}c). In contrast, applying the same steering vector with a negative multiplier leads the model to amplify perceived risk, generating more cautionary or morally disapproving language, such as ``wrong,'' ``never right,'' and ``not something I would ever do.'' Table \ref{tab:example_steered_generation} presents representative examples of text completions generated by the model under different steering conditions.

\setlength{\tabcolsep}{1pt}
\begin{table}[h!]
    \centering
    \caption{Example steered text completions generated by Gemma-2-9B-Instruct using a steering multiplier of 900. CA refers to Contrastive Activation, CE to Certainty Equivalent, and MCMC to Markov chain Monte Carlo with LLM.}
    \begin{tabular}{lccc} \toprule
        Risky event & \makecell{Steering\\method} & \makecell{Negative steer\\(risk averse)} & \makecell{Positive steer\\(risk seeking)} \\ \hline
        I think cheating on an exam... & CA & \makecell{is wrong because\\it undermines the\\integrity of the educational\\system and devalues\\the achievements of those\\who uphold ethical standards.} & is wrong.\\ \cline{2-4}
        & CE & \makecell{is absolutely wrong\\and absolutely not\\worth the consequences.} & is a weak move.\\ \cline{2-4}
        & MCMC & is wrong. & \makecell{is a very bad idea.\\I mean, sure, maybe\\you might get a slightly\\better grade, but in\\the long run, it's\\just going to compromise\\your learning.}\\ \cline{2-4}
        \makecell{I think stealing an additional\\TV cable connection off\\the one you pay for...} & CA & is unethical. &  is stealing. \\ \cline{2-4}
        & CE & \makecell{is not ethical and\\could have serious\\consequences.} & is slightly unethical. \\ \cline{2-4}
        & MCMC & is not ethical. & \makecell{is a slightly more minor\\offense compared to some\\other types of theft.} \\ \cline{2-4}
        \makecell{I think riding a motorcycle...} & CA & \makecell{is a great way to\\experience freedom and \\independence, especially for\\those who enjoy the outdoors\\and the wind in their hair.\\However, it's important\\to remember that motorcycling \\involves inherent risks.} & \makecell{is about more than just getting\\from point A to point B.\\It's about the feeling.} \\ \cline{2-4}
        & CE & \makecell{is absolutely amazing.\\The open air, the\\wind in your hair,\\the feeling of freedom\\and escape - it's\\truly unique.} & \makecell{is a rewarding but risky\\passtime. I enjoy the feeling\\of freedom and control\\I get while riding,\\but I also understand\\the dangers involved.} \\ \cline{2-4}
        & MCMC & \makecell{is not for everyone.\\It's not just about\\physical ability, it's also\\about mental and emotional\\preparedness.} &  \makecell{is a very exhilarating\\experience. I enjoy the\\feeling of the wind\\in my hair and the freedom\\of the open road.}\\ 
        \bottomrule
    \end{tabular}
    \label{tab:example_steered_generation}
\end{table}

\section{Discussion}

We investigated the use of aligned behavioral and neural representations of risk to steer LLM behavior across three risk-related domains: risky decision-making, risk perception, and text generation involving real-world risky events. In all three domains, steering vectors derived from aligned representations (i.e., the Certainty Equivalent and MCMC methods) consistently outperformed those generated via Contrastive Activation. These results suggest that self-alignment methods, which align the same latent construct elicited through both behavior and neural activations, offer a more effective and principled means of controlling LLM behavior in risk-sensitive contexts.

\textbf{Limitations and future research.}
While it is intuitive that self-aligned representations can be used to steer model behavior related to the underlying latent construct, further theoretical and mechanistic interpretability research is needed to establish a more concrete link between behaviorally elicited representations and their corresponding neural activations. Moreover, modifying a Transformer’s residual stream is not uniformly effective across all layers; the success of such interventions is highly dependent on the layer selected. Identifying optimal layers for steering remains an open question for future research. Finally, this form of steering is limited in scope for proprietary models where weights are not accessible, restricting its broader applicability.

\textbf{Acknowledgments.} This work and related results were made possible with the support of the NOMIS Foundation. HY acknowledges the Chancellor’s International Scholarship from the University of Warwick for additional support.

\bibliography{references}
\bibliographystyle{plain}








\appendix

\section{Prompts}

\subsection{Words for Contrastive Activation}
\label{ap:CA_words}
Risk words: \textit{`risk', `uncertainty', `danger', `volatility', `loss', `gamble', `exposure', `threat', `hazard', `insecurity', `unpredictability', `peril', `chance', `vulnerability', `instability', `jeopardy', `speculation', `probability', `accident', `daring'}

Safe words: \textit{`safety', `certainty', `stability', `gain', `assurance', `protection', `security', `safeguard', `reliability',`predictability', `refuge', `guarantee', `resilience', `steadiness', `shelter', `caution', `inevitability', `prevention', `prudence'}

\subsection{Prompts for Certainty Equivalent}

\textit{Starting with \$100 in capital, what is the maximum dollar amount you are willing to pay to participate in this gamble: the gamble offers a \{$p_1$\}\% chance to win \$100, a \{$p_2$\}\% chance to win \$50, a \{$p_3$\}\% chance to win \$0. Respond with a single numeric value only. Do not explain your reasoning. }

\subsection{Prompts for MCMC with LLM}
\textit{You are participating in a gambling game where you will be shown two options, Gamble A and Gamble B: }

\textit{Gamble A offers a \{$p_1$\}\% chance to win \$100, a \{$p_2$\}\% chance to win \$50, and a \{$p_3$\}\% chance to win \$0.}

\textit{Gamble B offers a \{$p_1'$\}\% chance to win \$100, a \{$p_2'$\}\% chance to win \$50, and a \{$p_3'$\}\% chance to win \$0. }

\textit{Your task is to choose between the two. Do not explain your reasoning, just limit your answer to either `A' or `B'.}

\subsection{Prompts for Neural Activations}
\label{ap:prompts_for_neural_acts}
\textit{You are offered a gambling game: the gamble offers a \{$p_1$\}\% chance to win \$100, a \{$p_2$\}\% chance to win \$50, a \{$p_3$\}\% chance to win \$0. Respond with a single word only to express how much does this gambling appeal to you. Do not explain your reasoning.}

\subsection{Prompts for Steering Risky Decision-Making}

\textit{Given the following gambles: Option A offers a 5\% chance to win \$100 and 95\% chance to win \$0; Option B offers a 100\% chance to win \$5. Please select either option A or option B to participate.}

\subsection{Prompts for Steering Risk Perception}

\textit{Please rate how risky this behavior is with a single numeric value ranging from 1 (Not at all) to 7 (Extremely risky): \{event\}}



\section{Comparing Steering Vectors}
\label{ap:compare_steer_vector}
\begin{figure}[h!]
    \centering
    \includegraphics[width=\linewidth]{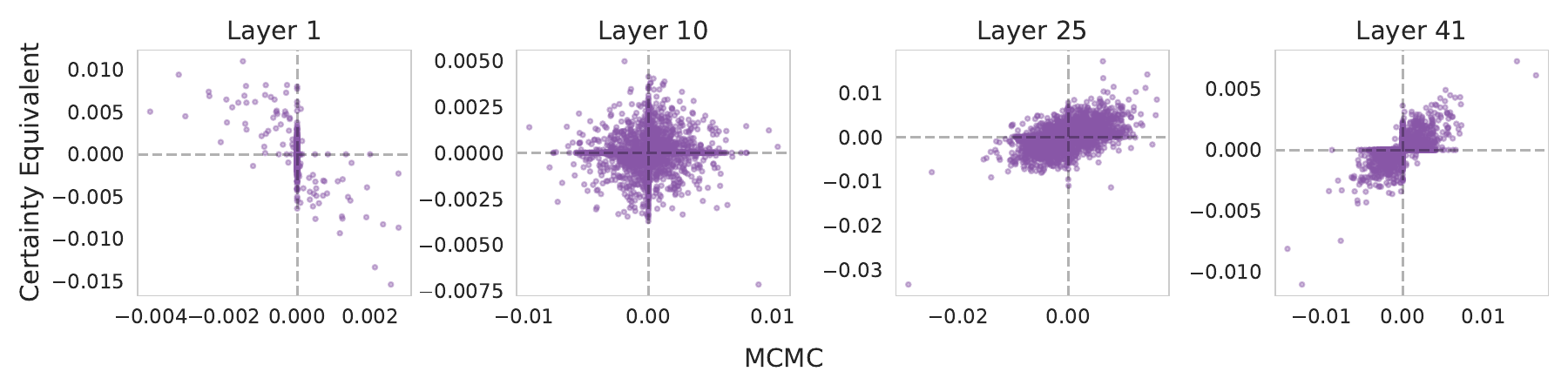}
    \caption{Comparison of steering vectors derived from MCMC with LLM (horizontal axis) and Certainty Equivalent (vertical axis) across selected layers of Gemma-2-9B-Instruct. Pearson correlation coefficients between the two steering vectors at layers 1, 10, 25, and 41 are $-0.63 ~(p<.01)$, $0.04~ (p=0.02)$, $0.48 ~(p<.01)$, and $0.64~ (p<.01)$, respectively. }
    \label{fig:compare_steer_vectors}
\end{figure}

In this section, we compare steering vectors derived from the MCMC and Certainty Equivalent methods across different layers of the Gemma-2-9B-Instruct model (see Figure~\ref{fig:compare_steer_vectors}). The only difference between the two approaches lies in the behavioral representation of risk used to compute alignment; the underlying neural representation remains identical. We observe a clear trend: as we move from earlier to later layers, the similarity between the two steering vectors increases, suggesting greater convergence in their influence on model behavior at deeper levels of information processing.

\section{Details of Representation Engineering}

In this section, we describe the procedure for injecting a steering vector, $h_A^l$, into the residual stream of a Transformer at layer $l$. At inference time, we first obtain the model’s neural activation at layer $l$ for a given prompt $p^*$:
\begin{align*}
    h^l \leftarrow M.\texttt{forward}(p^*).\texttt{activations}[l]
\end{align*}
where $M$ denotes the LLM and $h^l$ represents the original activation at layer $l$. Next, we modify this activation by injecting the steering vector scaled by a steering multiplier $c$:
\begin{align*}
    h_S^l \leftarrow h^l + ch_A^l
\end{align*}
where $h_S^l$ is the steered activation. The model then resumes its forward computation, beginning from layer $l$ with the modified activation:
\begin{align*}
    S \leftarrow M.\texttt{continue\_forward}(h_S^l)
\end{align*}
where $S$ denotes the final output generated by the steered model.

\section{Understanding the Marschak-Machina Triangle}

\begin{figure}[h!]
    \centering
    \includegraphics[width=0.9\linewidth]{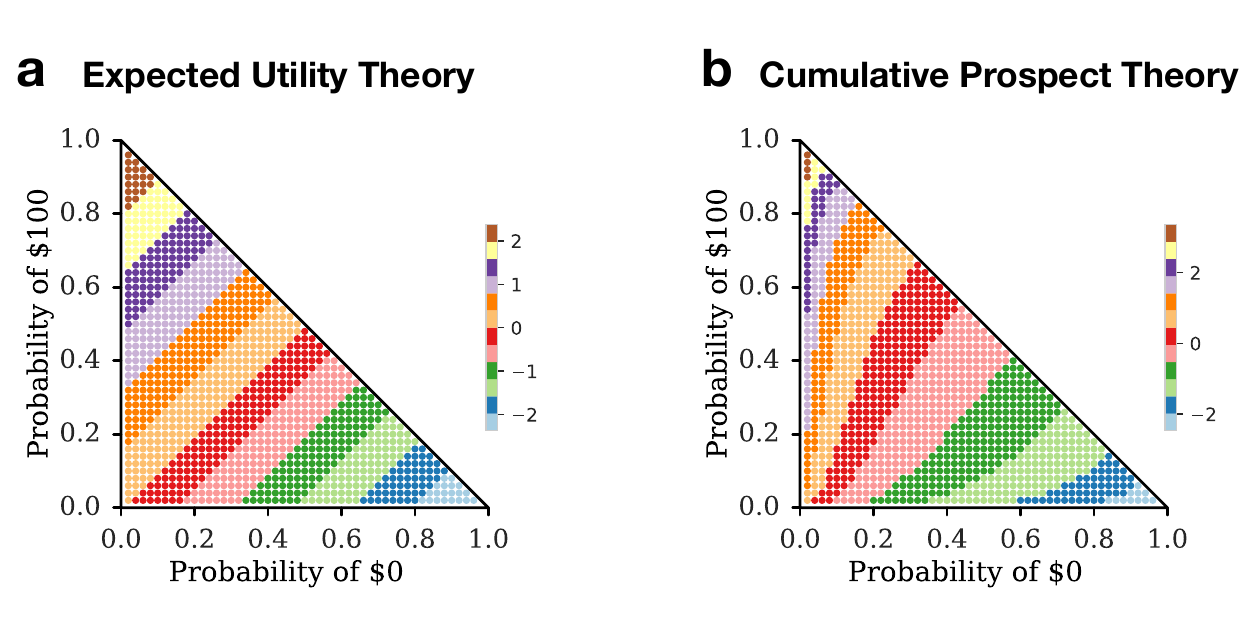}
    \caption{Predicted preferences under classical theories of human risky choice visualized in the Marschak–Machina probability triangle. \textbf{(a)} Expected utility theory \cite{von1947theory}. \textbf{(b)} Cumulative prospect theory \cite{tversky1992advances} using parameters $\alpha=0.88, \gamma=0.52$. In both panels, each point within the triangle represents a three-outcome gamble, and color intensity reflects the model’s predicted preference for that gamble. Higher values correspond to stronger preferences.}
    \label{fig:prob_triangle}
\end{figure}

As illustrated in Figure~\ref{fig:behavior_risk}, behavioral representations of risk are defined over the Marschak–Machina triangle \cite{marschak1950rational, machina1982expected}, a method used to characterize preferences over three-outcome gambles. To provide additional context, we visualize the theoretical predictions of two influential models of human risky choice within this triangle. Figure~\ref{fig:prob_triangle}a depicts predictions from Expected Utility Theory (EUT), a normative model of decision-making under risk \cite{von1947theory}, which assumes that individuals evaluate gambles based on the weighted sum of utility. Under EUT, indifference curves are always straight and parallel, reflecting consistent trade-offs between outcomes. In contrast, Cumulative Prospect Theory (CPT), a descriptive model that accounts for empirical deviations from EUT, generates indifference curves that exhibit a ``fanning out'' pattern (see Figure \ref{fig:prob_triangle}b). This curvature reflects increasing risk aversion as the probability of extreme outcomes changes, leading to steeper indifference curves in some regions of the triangle \cite{machina1982expected, harless1992predictions}.

Comparing the theoretical predictions within the probability triangle to the behavioral representation elicited via MCMC (Figure~\ref{fig:behavior_risk}b), we find that neither EUT nor CPT qualitatively captures the risk preferences exhibited by the Gemma-2-9B-Instruct model. This suggests that existing models of human risky choice may not be directly transferable to explaining LLM behavior \cite{liu2024large, zhu2024language}. Future research should consider developing new descriptive frameworks tailored to characterizing and predicting the risky choices of LLMs.

\section{Replication}
\label{ap:replication}

\begin{figure}[t!]
    \centering
    \includegraphics[width=0.9\linewidth]{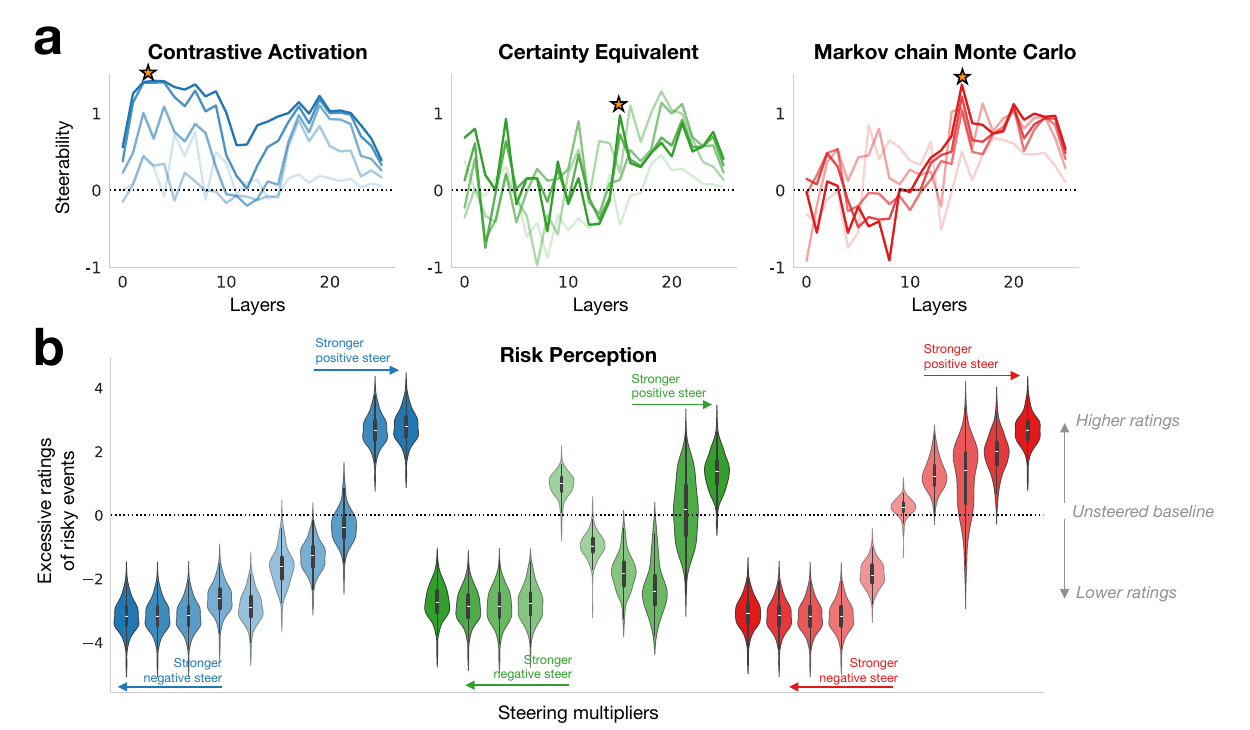}
    \caption{\textbf{Steering risk perception of Gemma-2-2B-Instruct.} \textbf{(a)} Steerability results using steering vectors derived from Contrastive Activation (blue), Certainty Equivalent (green), and MCMC (red). Darker colors represent larger steering multipliers. The optimal layers for steering, identified by the highest steerability at the maximum multiplier, are layers 3, 15, and 15 for the respective methods (marked with stars). \textbf{(b)} Change in average risk ratings for real-world events after steering, using the optimal layer for each method. The vertical axis reflects the deviation from the unsteered baseline rating. Each violin plot displays the distribution of ratings, with the white bar indicating the median and the black box representing the interquartile range up to the 75th percentile. }
    \label{fig:steer_risk_perception_2b}
\end{figure}

We replicated the main experiment using a smaller LLM: Gemma-2-2B-Instruct \cite{team2024gemma}.  

\textbf{Steering risky choices.}
For the four gambles presented in Table \ref{tab:four_fold_gambles}, Gemma-2-2B-Instruct exhibited a ceiling effect, consistently preferring the risky option with 100\% choice probability. As a result, all three steering vectors showed negligible steerability in this condition.

\textbf{Steering risk perception.}
Next, we examined the effects of steering vectors on the Gemma-2B model’s ratings of real-world risky events (see Figure \ref{fig:steer_risk_perception_2b}). The most steerable layer in the 2B model occurred at a similar relative depth as in the 9B model.

We conducted two separate two-way repeated-measures ANOVAs to evaluate the effects of steering method and multiplier on model ratings under positive and negative steering conditions.

For positive steering, the analysis revealed significant main effects of steering method, $F(2, 298) = 1710.77$, $p < .01$, and steering multiplier, $F(4, 596) = 5092.48$, $p < .01$, as well as a significant interaction between the two factors, $F(8, 1192) = 599.96$, $p < .01$. Follow-up paired $t$-tests showed that the MCMC method significantly outperformed both the Certainty Equivalent method ($t(149) = 55.63$, $p < .01$) and the Contrastive Activation method ($t(149) = 35.35$, $p < .01$). Additionally, the Contrastive Activation method outperformed the Certainty Equivalent method ($t(149) = 27.22$, $p < .01$).

For negative steering, we again found significant main effects of steering method, $F(2, 298) = 5786.59$, $p < .01$, and multiplier, $F(4, 596) = 2838.58$, $p < .01$, along with a significant interaction, $F(8, 1192) = 2640.75$, $p < .01$. Paired $t$-tests indicated that the MCMC method significantly outperformed the Certainty Equivalent method ($t(149) = 100.00$, $p < .01$) and the Contrastive Activation method ($t(149) = 11.47$, $p < .01$), while the Contrastive Activation method again outperformed the Certainty Equivalent method ($t(149) = 89.19$, $p < .01$).

\section{Implementation details}
Steered model completions were executed on a single A100 GPU, requiring approximately 50 hours for the 9B model and 40 hours for the 2B model. Computing steering vectors via self-alignment took an additional 2 hours on a single A100 GPU.


\end{document}